\documentclass[runningheads]{llncs}
\usepackage{graphicx}
\usepackage{amsmath} 
\usepackage{subcaption}
\begin{document}

\title{Understanding Individual Decisions of CNNs via Contrastive Backpropagation} 
\titlerunning{Contrastive LRP} 


\author{Jindong Gu$^{1,2}$ \and Yinchong Yang$^{2}$ \and Volker Tresp$^{1,2}$}

%

\authorrunning{Jindong Gu et al.} 


\institute{The University of Munich, Munich, Germany \\ \and
Siemens AG, Corporate Technology, Munich, Germany \\ }

\maketitle

\begin{abstract}
A number of backpropagation-based approaches such as DeConvNets, vanilla Gradient Visualization and Guided Backpropagation have been proposed to better understand individual decisions of deep convolutional neural networks. The saliency maps produced by them are proven to be non-discriminative. Recently, the Layer-wise Relevance Propagation (LRP) approach was proposed to explain the classification decisions of rectifier neural networks. In this work, we evaluate the discriminativeness of the generated explanations and analyze the theoretical foundation of LRP, i.e. Deep Taylor Decomposition. The experiments and analysis conclude that the explanations generated by LRP are not class-discriminative. Based on LRP, we propose Contrastive Layer-wise Relevance Propagation (CLRP), which is capable of producing instance-specific, class-discriminative, pixel-wise explanations. In the experiments, we use the CLRP to explain the decisions and understand the difference between neurons in individual classification decisions. We also evaluate the explanations quantitatively with a Pointing Game and an ablation study. Both qualitative and quantitative evaluations show that the CLRP generates better explanations than the LRP. The code is available \footnote{https://github.com/Jindong-Explainable-AI/Contrastive-LRP}.
\end{abstract}
\section{Introduction}
Deep convolutional neural networks (DCNNs) achieve start-of-the-art performance on many tasks, such as visual object recognition\cite{krizhevsky2012imagenet,simonyan2014very,szegedy2015going}, and object detection\cite{girshick2014rich,redmon2016you}. However, since they lack transparency, they are considered as "black box" solutions. Recently, research on explainable deep learning has received increased attention:  Many approaches have been proposed to crack the "black box". Some of them aim to interpret the components of a deep-architecture model and understand the image representations extracted from deep convolutional architectures \cite{mahendran2015understanding,dosovitskiy2016inverting,mahendran2016visualizing}. Examples are Activation Maximization \cite{erhan2009visualizing,simonyan2013deep}, DeConvNets Visualization \cite{zeiler2014visualizing}.
Others focus on explaining the individual classification decisions. Examples are Prediction Difference Analysis \cite{robnik2008explaining,shrikumar2017learning}, Guided Backpropagation \cite{simonyan2013deep,springenberg2014striving}, Layer-wise Relevance Propagation (LRP) \cite{bach2015pixel,montavon2017explaining}, Class Activation Mapping \cite{zhou2016learning,selvaraju2016grad} and Local Interpretable Model-agnostic Explanations \cite{ribeiro2016should,ribeiro2016nothing}.

More concretely, the models in \cite{oquab2015object,zhou2016learning} were originally proposed to detect object only using category labels. They work by producing saliency maps of objects corresponding to the category labels. Their produced saliency maps can also explain the classification decisions to some degree. However, the approaches can only work on the model with a specific architecture. For instance, they might require a fully convolutional layer followed by a max-pooling layer, a global average pooling layer or an aggregation layer, before a final softmax output layer. The requirement is not held in most off-the-shelf models e.g., in \cite{krizhevsky2012imagenet,simonyan2014very}. The perturbation methods \cite{ribeiro2016should,ribeiro2016nothing,robnik2008explaining} require no specific architecture. For a single input image, however, they require many instances of forward inference to find the corresponding classification explanation, which is computationally expensive.

The backpropagation-based approaches \cite{simonyan2013deep,springenberg2014striving,bach2015pixel} propagate a signal from the output neuron backward through the layers to the input space in a single pass, which is computationally efficient compared to the perturbation methods. They can also be applied to the off-the-shelf models. In this paper, we focus on the backpropagation approaches. The outputs of the backpropagation approaches are instance-specific because these approaches leverage the instance-specific structure information (ISSInfo). The ISSInfo, equivalent to \textit{bottleneck} information in \cite{mahendran2016salient}, consist of selected information extracted by the forward inference, i.e., the Pooling switches and ReLU masks. With the ISSInfo, the backpropagation approaches can generate instance-specific explanations. A note on terminology: although the terms "sensitivity map", "saliency map", "pixel attribution map" and "explanation heatmap" may have different meanings in different contexts, in this paper, we do not distinguish them and use the term "saliency map" and "explanation" interchangeably.

The primal backpropagation-based approaches, e.g., the vanilla Gradient Visualization \cite{simonyan2013deep} and the Guided Backpropagation \cite{springenberg2014striving} are proven to be inappropriate to study the neurons of networks because they produce non-discriminative saliency maps \cite{mahendran2016salient}. The saliency maps generated by them mainly depend on ISSInfo instead of the neuron-specific information. In other words, the generated saliency maps are not class-discriminative with respect to class-specific neurons in output layer. The saliency maps are selective of any recognizable foreground object in the image \cite{mahendran2016salient}. Furthermore, the approaches cannot be applied to understand neurons in intermediate layers of DCNNs, either. In \cite{zeiler2014visualizing,gonzalez2018semantic}, the differences between neurons of an intermediate layer are demonstrated by a large dataset. The neurons are often activated by certain specific patterns. However, the difference between single neurons in an individual classification decision has not been explored yet. In this paper, we will also shed new light on this topic. 

The recently proposed Layer-wise Relevance Propagation (LRP) approach is proven to outperform the gradient-based approaches \cite{montavon2017explaining}. Apart from explaining image classifications\cite{lapuschkin2017understanding,montavon2017explaining}, the LRP is also applied to explain the classifications and predictions in other tasks \cite{srinivasan2017interpretable,arras2017explaining}. However, the explanations generated by the approach has not been fully verified. We summarise our three-fold contributions as follows:
\begin{itemize}
\item[1] We first evaluate the explanations generated by LRP for individual classification decisions. Then, we analyze the theoretical foundation of LRP, i.e., Deep Taylor Decomposition and shed new insight on LRP.
\item[2] We propose Contrastive Layer-wise Relevance Propagation (CLRP). To generate class-discriminative explanations, we propose two ways to model the contrastive signal (i.e., an opposite visual concept). For individual classification decisions, we illustrate explanations of the decisions and the difference between neuron activations using the proposed approach.
\item[3] We build a GPU implementation of LRP and CLRP using Pytorch Framework, which alleviates the inefficiency problem addressed in \cite{zhang2016top,shrikumar2017learning}.
\end{itemize}

Related work is reviewed in the next section. Section \ref{sec:rethink_LRP} analyzes LRP theoretically and experimentally. In Section \ref{sec:const_backp}, the proposed approach CLRP is introduced. Section \ref{sec:experiments} shows experimental results to evaluate the CLRP qualitatively and quantitatively on two tasks, namely, explaining the image classification decisions and understanding the difference of neuron activations in single forward inference. The last section contains conclusions and discusses future work.

\section{Related Work}
The DeConvNets were originally proposed for unsupervised feature learning tasks \cite{zeiler2010deconvolutional}. Later they were applied to visualize units in convolutional networks \cite{zeiler2014visualizing}. The DeConvNets maps the feature activity to input space using ISSInfo and the weight parameters of the forward pass. \cite{simonyan2013deep} proposed identifying the vanilla gradients of the output with respect to input variables are their relevance. The work also showed its relation to the DeConvNets. They use the ISSInfo in the same way except for the handling of rectified linear units (ReLUs) activation function. The Guided Backpropagation \cite{springenberg2014striving} combine the two approaches to visualize the units in higher layers.

The paper \cite{bach2015pixel} propose LRP to generate the explanations for classification decisions. The LRP propagates the class-specific score layer by layer until to input space. The different propagation rules are applied according to the domain of the activation values. \cite{montavon2017explaining} proved that the Taylor Expansions of the function at the different points result in the different propagation rules. Recently, one of the propagation rules in LRP, \textit{z}-rule, has been proven to be equivalent to the vanilla gradients (saliency map in \cite{simonyan2013deep}) multiplied elementwise with the input \cite{kindermans2016investigating}. The vanilla Gradient Visualization and the Guided Backpropagation are shown to be not class-discriminative in \cite{mahendran2016salient}. This paper rethinks the LRP and evaluates the explanations generated by the approach.

Existing work that is based on discriminative and pixel-wise explanations are \cite{cao2015look,zhang2016top,selvaraju2016grad}. The work Guided-CAM \cite{selvaraju2016grad} combines the low-resolution map of CAM and the pixel-wise map of Guided Backpropagation to generate a pixel-wise and class-discriminative explanation. To localize the most relevant neurons in the network, a biologically inspired attention model is proposed in \cite{tsotsos1995modeling}. The work uses a top-down (from the output layer to the intermediate layers) Winner-Take-All process to generate binary attention maps. The work \cite{zhang2016top} formulate the top-down attention of a CNN classifier as a probabilistic Winner-Take-All process. The work also uses a contrastive top-down attention formulation to enhance the discriminativeness of the attention maps. Based on their work and the LRP, we propose Contrastive Layer-wise Relevance Propagation (CLRP) to produce class-discriminative and pixel-wise explanations. Another publication related to our approach is \cite{cao2015look}, which is able to produce class-discriminative attention maps. While the work \cite{cao2015look} requires modifying the traditional CNNs by adding extra feedback layers and optimizing the layers during the backpropagation, our proposed methods can be applied to all exiting CNNs without any modification and further optimization.

\section{Rethinking Layer-wise Relevance Propagation}
\label{sec:rethink_LRP}
Each neuron in DCNNs represents a nonliear function $X^{L+1}_i = \phi(\boldsymbol{X}^L \boldsymbol{W}^L_i+\boldsymbol{b}^L_{i})$, where $\phi$ is an activation function and $\boldsymbol{b}^L_{i}$ is a bias for the neuron $X^{L+1}_i$. The inputs of the nonliear function corresponding to a neuron are the activation values of the previous layer $\boldsymbol{X}_{i}$ or the raw input of the network. The output of the function are the activation values of the neuron $X^{L+1}_i$. The whole network are composed of the nested nonlinear functions.

To identify the relevance of each input variables, the LRP propagates the activation value from a single class-specific neuron back into the input space, layer by layer. The logit before softmax normalization is taken, as explained in \cite{simonyan2013deep,bach2015pixel}. In each layer of the backward pass, given the relevance score $\boldsymbol{R}^{L+1}$ of the neurons $\boldsymbol{X}^{L+1}$, the relevance $R^{L}_{i}$ of the neuron $X^{L}_{i}$ are computed by redistributing the relevance score using local redistribution rules. The most often used rules are the $z^+$-rule and the $z^\beta$-rule, which are defined as follows:
\begin{equation}
\begin{split}
\text{$z^+$-rule:\ } R^{L}_i &=  \sum_{j} \frac{x_i w^+_{ij}}{\sum_{i^\prime} x_{i^\prime} w^+_{{i^\prime}j}} R^{L+1}_j \\
\text{$z^\beta$-rule:\ }  R^{L}_i &=  \sum_{j} \frac{x_i w_{ij} -l_i w^+_{ij} - h_i w^-_{ij}}{\sum_{i^\prime} x_{i^\prime} w_{{i^\prime}j} -l_{i^\prime} w^+_{{i^\prime}j} - h_{i^\prime} w^-_{{i^\prime}j}} R^{L+1}_j
\end{split}
\end{equation}
where $w_{ij}$ connecting $X^{L}_{i}$ and $X^{L+1}_{j}$ is a parameter in $L$-th layer, $w^+_{ij} = w_{ij}*1_{w_{ij}>0}$ and $w^-_{ij} = w_{ij}*1_{w_{ij}<0}$, and the interval $[l, h]$ is the domain of the activation value $x_i$.

\subsection{Evaluation of the Explanations Generated by the LRP}
\label{sec:lrp_eval}

\begin{figure}
    \centering
    \begin{subfigure}{\textwidth}
        \centering
        \includegraphics[scale=0.28]{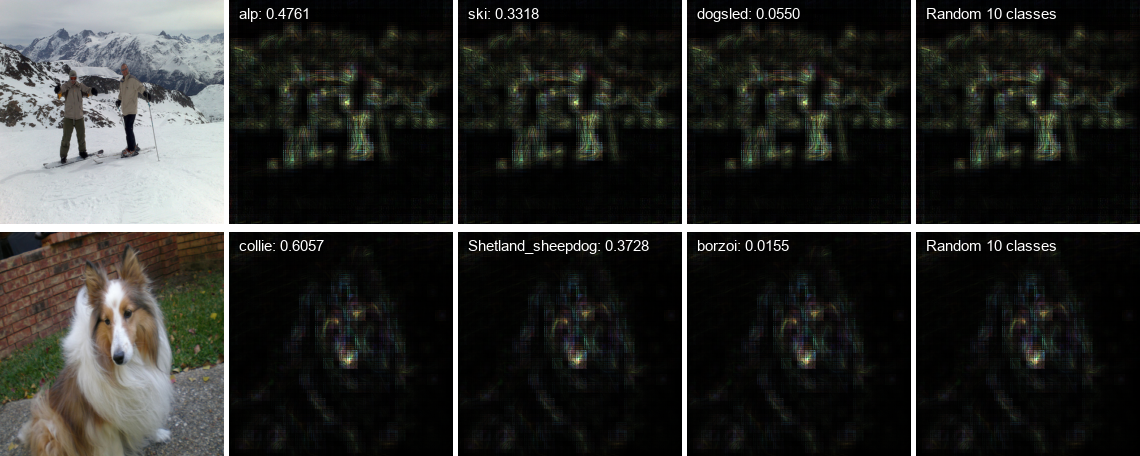}
        \caption{The explanations generated by LRP on AlexNet.}
    \end{subfigure}
    \vspace{10pt}

     \begin{subfigure}{\textwidth}
        \centering
        \includegraphics[scale=0.28]{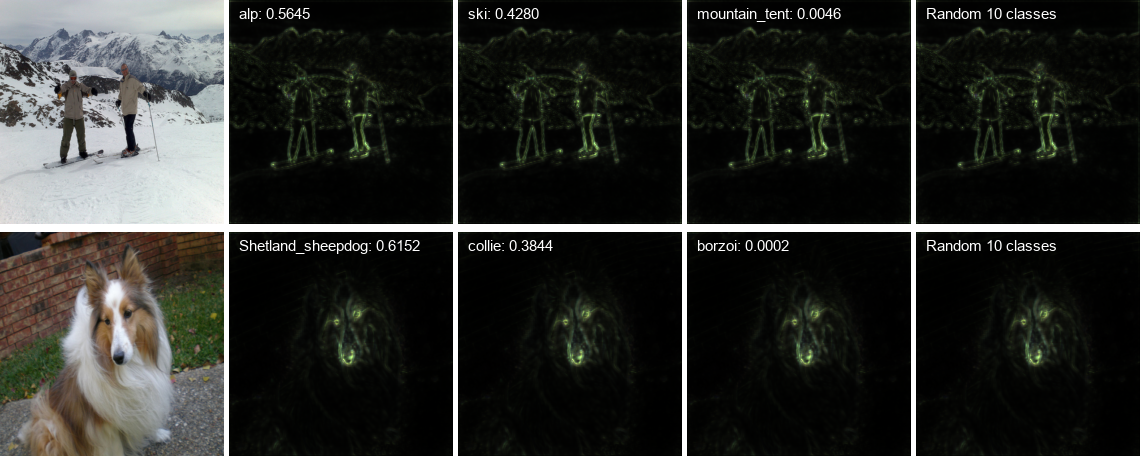}
        \caption{The explanations generated by LRP on VGG16 Network.}
    \end{subfigure}
    \vspace{10pt}

    \begin{subfigure}{\textwidth}
        \centering
        \includegraphics[scale=0.23]{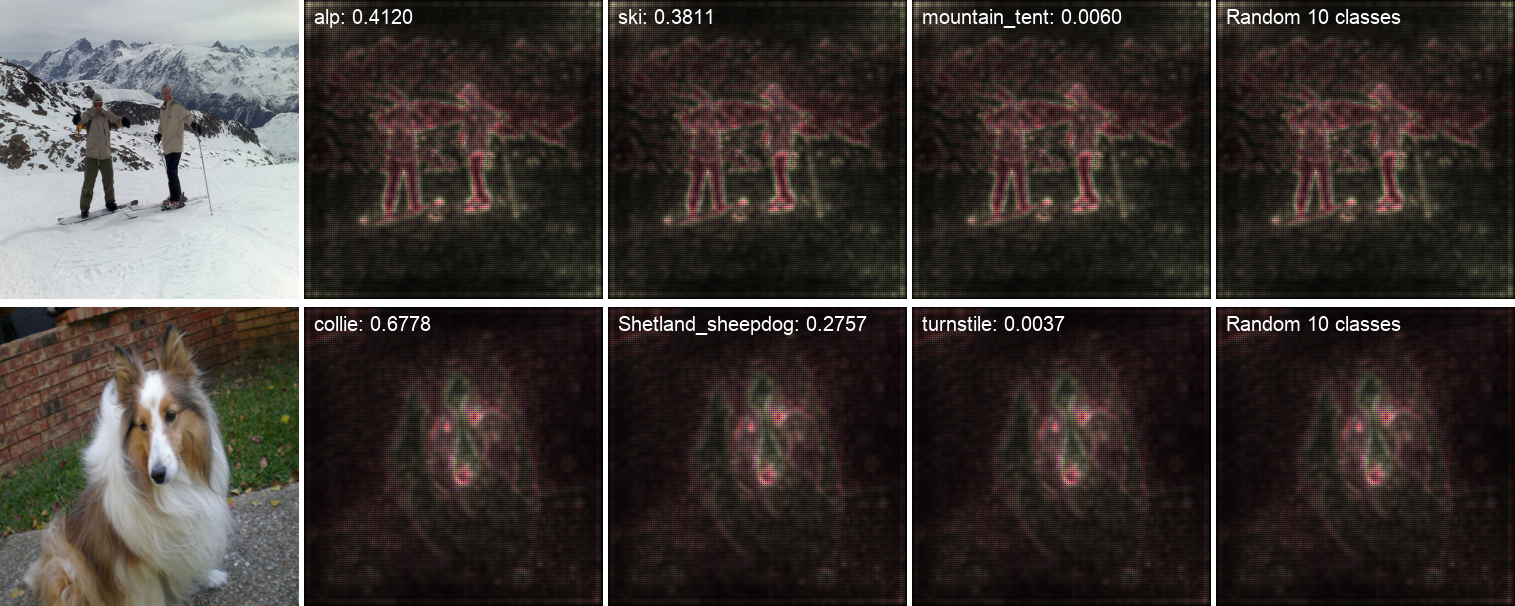}
        \caption{The explanations generated by LRP on GoogLeNet.}
        \label{fig:lrp_eval_google}
    \end{subfigure}
    \caption{The images from validation datasets of ImageNet are classified using the off-the-shelf models pre-trained on the ImageNet. The classifications of the images are explained by the LRP approach. For each image, we generate four explanations that correspond to the top-3 predicted classes and a randomly chosen multiple-classes.}
\label{fig:lrp_eval}
\end{figure}

The explanations generated by LRP are known to be instance-specific. However, the discriminativeness of the explanations has not been evaluated yet.  Ideally, the visualized objects in the explanation should correspond to the class the class-specific neuron represents. We evaluate the explanations generated by LRP on the off-the-shelf models from $torchvision$, specifically, AlexNet \cite{krizhevsky2012imagenet}, VGG16 \cite{simonyan2014very} and GoogLeNet \cite{szegedy2015going} pre-trained on the ImageNet dataset \cite{russakovsky2015imagenet}.

The experiment settings are similar to \cite{montavon2017explaining}. The $z^\beta$-rule is applied to the first convolution layer. For all higher convolutional layers and fully-connected layers, the $z^+$-rule is applied. In the MaxPooling layers, the relevance is only redistributed to the neuron with the maximal value inside the pooling region, while it is redistributed evenly to the corresponding neurons in the Average Pooling layers. The biases and normalization layers are bypassed in the relevance propagation pass.

The results are shown in figure \ref{fig:lrp_eval}. For each test image, we create four saliency maps as explanations. The first three explanation maps are generated for top-3 predictions, respectively. The fourth one is created for randomly chosen 10 classes from the top-100 predicted classes (which ensure that the score to be propagated is positive). The white text in each explanation map indicates the class the output neuron represents and the corresponding classification probability. The explanations generated by AlexNet are blurry due to incomplete learning (due to the limited expressive power). The explanations of VGG16 classifications are sharper than the ones created on GoogLenet. The reason is that VGG16 contains only MaxPooling layers and GoogLenet, by contrast, contains a few average pooling layers.

The generated explanations are instance-specific, but not class-discriminative. In other words, they are independent of class information. The explanations for different target classes, even randomly chosen classes, are almost identical. The conclusion is consistent with the one summarised in the paper \cite{dosovitskiy2016inverting,agrawal2014analyzing}, namely, almost all information about input image is contained in the pattern of non-zero pattern activations, not their precise values. The high similarity of those explanations resulted from the leverage of the same ISSInfo (see section \ref{fig:foundation}). In summary, the explanations are not class-discriminative. The generated maps recognize the same foreground objects instead of a class-discriminative one.

\subsection{Theoretical Foundation: Deep Taylor Decomposition}
\label{fig:foundation}
Motivated by the divide-and-conquer paradigm, Deep Taylor Decomposition decomposes a deep neural network (i.e. the nested nonliear functions) iteratively \cite{montavon2017explaining}. The propagation rules of LRP are derivated from Deep Taylor Decomposition of rectifier neuron network. The function represented by a single neuron is $X^{L+1}_j = max(0, \boldsymbol{X}^L \boldsymbol{W}^L_j + b^{L+1}_j)$. The relevance $R^{L+1}_{j}$ of the neurons $X^{L+1}_j$ is given. The Deep Taylor Decomposition assumes $R^{L+1}_{j} = max(0, \boldsymbol{X}^L \boldsymbol{W}^L_j + b^{L+1}_j)$. The function is expanded with Taylor Series at a point $\boldsymbol{X}_i^r$ subjective to $max(0, \boldsymbol{X^r}^L \boldsymbol{W}^L_j + b^{L+1}_j) = 0$. The LRP propagation rules are resulted from the first degree terms of the expansion.

One may hypothesize that the non-discriminativeness of LRP is caused by the first-order approximation error in Deep Taylor Decomposition. We proved that, under the given assumption, the same propagation rules are derived, even though all higher-order terms are taken into consideration (see the proof in the supplementary material). Furthermore, we found that the theoretical foundation provided by the Deep Taylor Decomposition is inappropriate. The assumption $R^{L+1}_{j} = max(0, \boldsymbol{X}^L \boldsymbol{W}^L_j + b^{L+1}_j) = X^{L+1}_j$ is not held at all the layers except for the last layer. The assumption indicates that the relevance value is equal to the activation value for all the neurons, which, we argue, is not true.

In our opinion, the explanations generated by the LRP result from the ISSInfo (ReLU masks and Pooling Switches). The activation values of neurons are required to create explanations using LRP. In the forward pass, the network output a vector ($y_1, y_2, \cdots, y_m$). In the backward pass, the activation value of the class $y_1$ is layer-wise backpropagated into input space. In fully connected layers, only the activated neurons can receive the relevance according to any LRP propagation rule. In the Maxpooling layers, the backpropagation conducts an unpooling process, where only the neuron with maximal activations inside the corresponding pooling region can receive relevance. In the convolutional layer, only specific part of neurons $R_{conv1}$ in feature map have non-zero relevance in the backward pass. The part of input pixels $P_{input}$ live in the convolutional regions of those neurons $(R_{conv1})$. Only the pixels $P_{input}$ will receive the propagated relevance. The pattern of the $P_{input}$ is the explanation generated by LRP. 

The backward pass for the class $y_2$ is similar to that of $y_1$. The neurons that receive non-zero relevance are the same as in case of $y_1$, even though their absolute values may be slightly different. Regardless of the class chosen for the backpropagation, the neurons of each layer that receive non-zero relevance stay always the same. In other words, the explanations generated by LRP are independent of the class category information, i.e., not class-discriminative.

In summary, in deep convolutional rectifier neuron network, the ReLU masks and Pooling Switches decide the pattern visualized in the explanation, which is independent of class information. That is the reason why the explanations generated by LRP on DCNNs are not class-discriminative. The analysis also explains the non-discriminative explanations generated by other backpropagation approaches, such as the DeConvNets Visualization \cite{zeiler2014visualizing}, The vanilla Gradient Visualization \cite{simonyan2013deep} and the Guided Backpropagation \cite{springenberg2014striving}. 

\section{Contrastive Layer-wise Relevance Propagation}
\label{sec:const_backp}
Before introducing our CLRP, we first discuss the conservative property in the LRP. In a DNN, given the input $\boldsymbol{X} = \{x_1, x_2, x_3,\cdots, x_n\}$, the output $\boldsymbol{Y} = \{y_1, y_2, y_3,\cdots, y_m\}$, the score $S_{y_j}$ (activation value) of the neuron $y_j$ before softmax layer, the LRP generate an explanation for the class $y_j$ by redistributing the score $S_{y_j}$ layer-wise back to the input space. The assigned relevance values of the input neurons are $\boldsymbol{R} = \{r_1, r_2, r_3,\cdots, r_n\}$. The conservative property is defined as follows:

\begin{definition}
The generated saliency map is conservative if the sum of assigned relevance values of the input neurons is equal to the score of the class-specific neuron, $\sum_{i=1}^{n} r_i=S_{y_j}$.
\end{definition}
\begin{figure}[t]
    \centering
        \centering
        \includegraphics[scale=0.22]{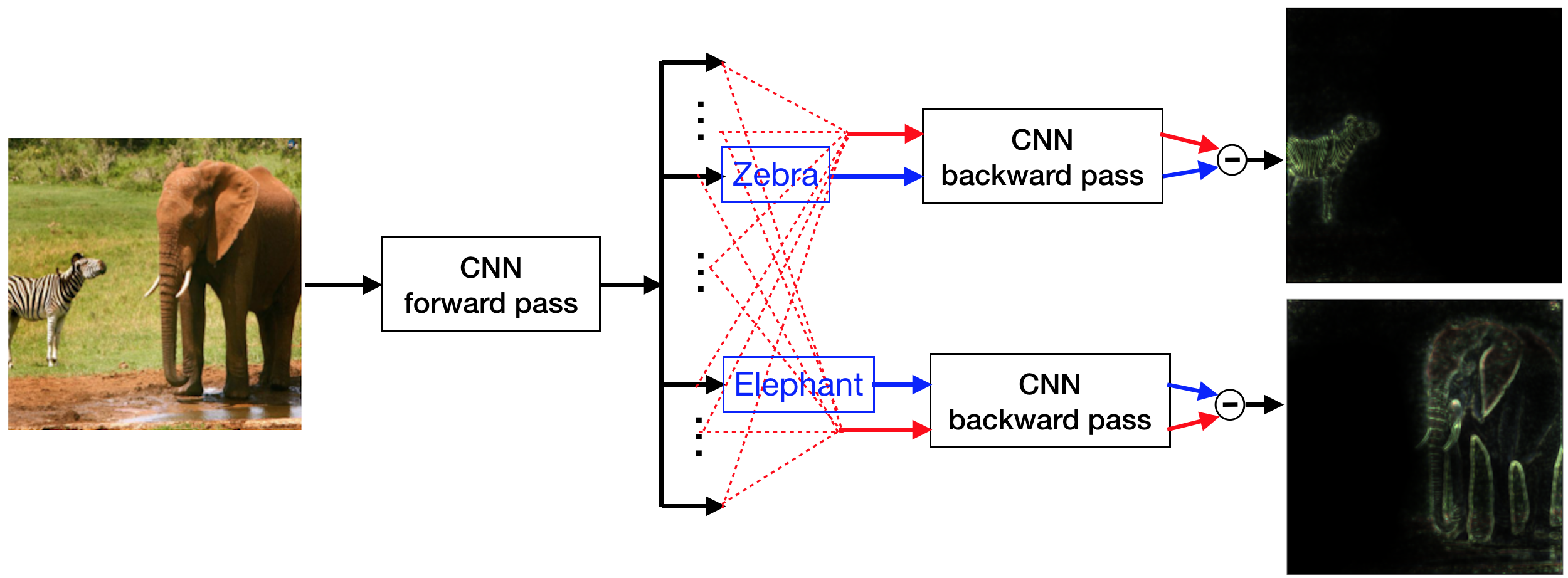}
    \caption{The figure shows an overview of our CLRP. For each predicted class, the approach generates a class-discriminative explanation by comparing two signals. The blue line means the signal that the predicted class represents. The red line models a dual concept opposite to the predicted class. The final explanation is the difference between the two saliency maps that the two signal generate.}
\label{fig:overview_clrp}
\end{figure}

In this section, we consider redistributing the same score from different class-specific neurons respectively. The assigned relevance $\boldsymbol{R}$ are different due to different weight connections. However, the non-zero patterns of those relevance vectors are almost identical, which is why LRP generate almost the same explanations for different classes. The sum of each relevance vector is equal to the redistributed score according to the conservative property. The input variables that are discriminative to each target class are a subset of input neurons, i.e., $\boldsymbol{X}_{dis} \subset \boldsymbol{X}$. The challenge of producing the explanation is to identify the discriminative pixels $\boldsymbol{X}_{dis}$ for the corresponding class.

In the explanations of image classification, the pixels on salient edges always receive higher relevance value than other pixels including all or part of $\boldsymbol{X}_{dis}$. Those pixels with high relevance values are not necessary discriminative to the corresponding target class. We observe that $\boldsymbol{X}_{dis}$ receive higher relevance values than that of the same pixels in explanations for other classes. In other words, we can identify $\boldsymbol{X}_{dis}$ by comparing two explanations of two classes. One of the classes is the target class to be explained. The other class is selected as an auxiliary to identify $\boldsymbol{X}_{dis}$ of the target class. To identify $\boldsymbol{X}_{dis}$ more accurately, we construct a virtual class instead of selecting another class from the output layer. We propose two ways to construct the virtual class.

The overview of the CLRP are shown in figure \ref{fig:overview_clrp}. We describe the CLRP formally as follows. The $j-$th class-specific neuron $y_{j}$  is connected to input variables by the weights $\boldsymbol{W} = \{\boldsymbol{W}^1, \boldsymbol{W}^2, \cdots, \boldsymbol{W}^{L-1}, \boldsymbol{W}^L_{j}\}$ of layers between them, where $\boldsymbol{W}^L$ means the weights connecting the $(L-1)-$th layer and the $L-$th layer, and $\boldsymbol{W}^L_{j}$ means the weights connecting the $(L-1)-$th layer and the $j-$th neuron in the $L-$th layer. The neuron $y_{j}$ models a visual concept $O$. For an input example $\boldsymbol{X}$, the LRP maps the score $S_{y_{j}}$ of the neuron back into the input space to get relevance vector $\boldsymbol{R} = f_{LRP} (\boldsymbol{X}, \boldsymbol{W}, S_{y_{j}})$.

We construct a dual virtual concept $\overline{O}$, which models the opposite visual concept to the concept $O$. For instance, the concept $O$ models the \textbf{zebra}, and the constructed dual concept $\overline{O}$ models the \textbf{non-zebra}. One way to model the $\overline{O}$ is to select all classes except for the target class representing $O$. The concept $\overline{O}$ is represented by the selected classes with weights $\overline{\boldsymbol{W}} = \{\boldsymbol{W}^1, \boldsymbol{W}^2, \cdots, \boldsymbol{W}^{L-1}, \boldsymbol{W}^L_{\{-j\}}\}$, where $\boldsymbol{W}_{\{-j\}}$ means the weights connected to the output layer excluding the $j-$th neuron. E.g. the dashed red lines in figure \ref{fig:overview_clrp} are connected to all classes except for the target class \textbf{zebra}. Next, the score $S_{y_{j}}$ of target class is uniformly redistributted to other classes. Given the same input example $\boldsymbol{X}$, the LRP generates an explanation $\boldsymbol{R}_{dual} = f_{LRP}(\boldsymbol{X}, \overline{\boldsymbol{W}}, S_{y_{j}})$ for the dual concept. The Contrastive Layer-wise Relevance Propagation is defined as follows:
\begin{equation}
\boldsymbol{R}_{CLRP} =\max(\boldsymbol{0}, (\boldsymbol{R} - \boldsymbol{R}_{dual}))
\end{equation}
where the function $\max(\boldsymbol{0}, \boldsymbol{X})$ means replacing the negative elements of $\boldsymbol{X}$ with zeros. The difference between the two saliency maps cancels the common parts. Without the dominant common parts, the non-zero elements in $\boldsymbol{R}_{CLRP}$ are the most relevant pixels $\boldsymbol{X}_{dis}$. If the neuron $y_{j}$ lives in an intermediate layer of a neural network, the constructed $\boldsymbol{R}_{CLRP}$ can be used to understand the role of the neuron.

Similar to \cite{zhang2016top}, the other way to model the concept $\overline{O}$ is to negate the weights $W_{ij}$. The concept $\overline{O}$ can be represented by the weights $\overline{\boldsymbol{W}} = \{\boldsymbol{W}^1, \boldsymbol{W}^2, \cdots, \boldsymbol{W}^{L -1},$ $ -1*\boldsymbol{W}^L_{j}\}$. All the weights are same as in the concept $O$ except that the weights of the last layer $\boldsymbol{W}^L_{j}$ are negated. In the experiments section, we call the first modeling method CLRP1 and the second one CLRP2. The contrastive formulation in \cite{zhang2016top} can be applied to other backpropagation approaches by normalizing and subtracting two generated saliency maps. However, the normalization strongly depends on the maximal value that could be caused by a noisy pixel. Based on the conservative property of LRP, the normalization is avoided in the proposed CLRP.

\section{Experiments and Analysis}
\label{sec:experiments}
In this section, we conduct experiments to evaluate our proposed approach. The first experiment aims to generate class-discriminative explanations for individual classification decisions. The second experiment evaluates the generated explanations quantitatively on the ILSVRC2012 validation dataset. The discriminativeness of the generated explanations is evaluated via a Pointing Game and an ablation study. The last experiment aims to understand the difference between neurons in a single classification forward pass.

\subsection{Explaining Classification Decisions of DNNs}
\begin{figure}[t]
    \centering
        \centering
        \includegraphics[scale=0.20]{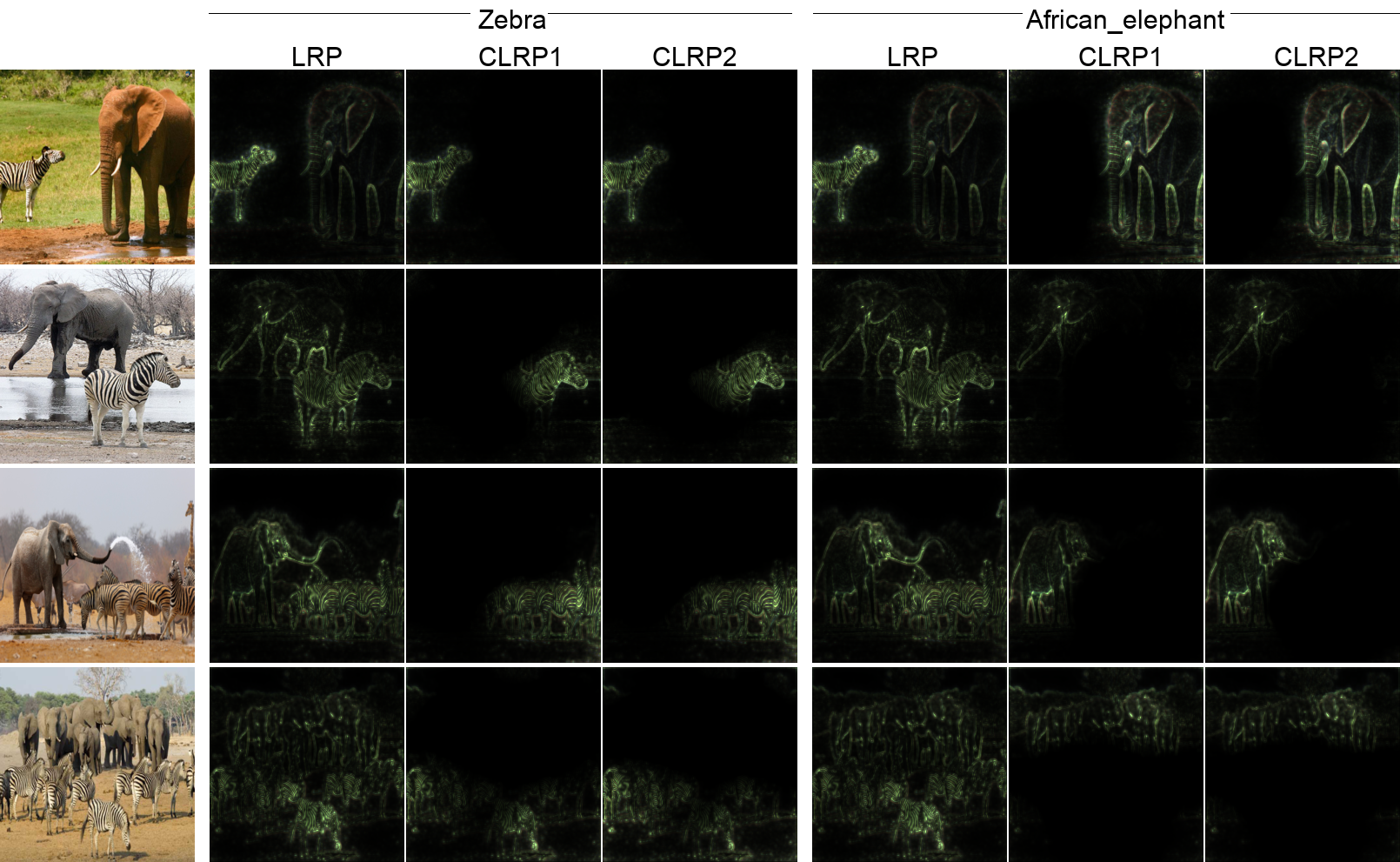}
    \caption{The images of multiple objects are classified using VGG16 network pre-trained on ImageNet. The explanations for the two relevant classes are generated by LRP and CLRP. The CLRP generates class-discriminative explanations, while LRP generates almost same explanations.}
\label{fig:vgg_exp}
\end{figure}
In this experiment, the LRP, the CLRP1 and the CLRP2 are applied to generate explanations for different classes. The experiments are conducted on a pre-trained VGG16 Network \cite{simonyan2014very}. The propagation rules used in each layer are the same as in the section \ref{sec:lrp_eval}. We classify the images of multiple objects. The explanations are generated for the two most relevant predicted classes, respectively. The figure \ref{fig:vgg_exp} shows the explanations for the two classes (i.e., $Zebra$ and $African\_elephant$). The explanations generated by the LRP are same for the two classes. Each generated explanation visualizes both $Zebra$ and $African\_elephant$, which is not class-discriminative. By contrast, both CLRP1 and CLRP2 only identify the discriminative pixels related to the corresponding class. For the target class $Zebra$, only the pixels on the zebra object are visualized. Even for the complicated images where a zebra herd and an elephant herd co-exist, the CLRP methods are still able to find the class-discriminative pixels.

We evaluate the approach with a large number of images with multiple objects. The explanations generated by CLRP are always class-discriminative, but not necessarily semantically meaningful for every class. One of the reasons is that the VGG16 Network is not trained for multi-label classification. Other reasons could be the incomplete learning and bias in the training dataset \cite{torralba2011unbiased}.

The implementation of the LRP is not trivial. The one provided by their authors only supports CPU computation. For the VGG16 network, it takes the 30s to generate one explanation on an Intel Xeon 2.90GHz $\times$ 6 machine. The computational expense makes the evaluation of LRP impossible on a large dataset \cite{zhang2016top}. We implement a GPU version of the LRP approach, which reduces the 30s to 0.1824s to generate one explanation on a single NVIDIA Telsa K80 GPU. The implementation alleviates the inefficiency problem addressed in \cite{zhang2016top,shrikumar2017learning} and makes the quantitative evaluation of LRP on a larget dataset possible.

\subsection{Evaluating the explanations}
In this experiments, we quantitatively evaluate the generated explanations on the  ILSVRC2012 validation dataset containing 50, 000 images. A Pointing Game and an ablation study are used to evaluate the proposed approach.

\begin{figure}[t]
    \centering
    \begin{subfigure}[t]{0.48\textwidth}
        \centering
        \includegraphics[scale=0.41]{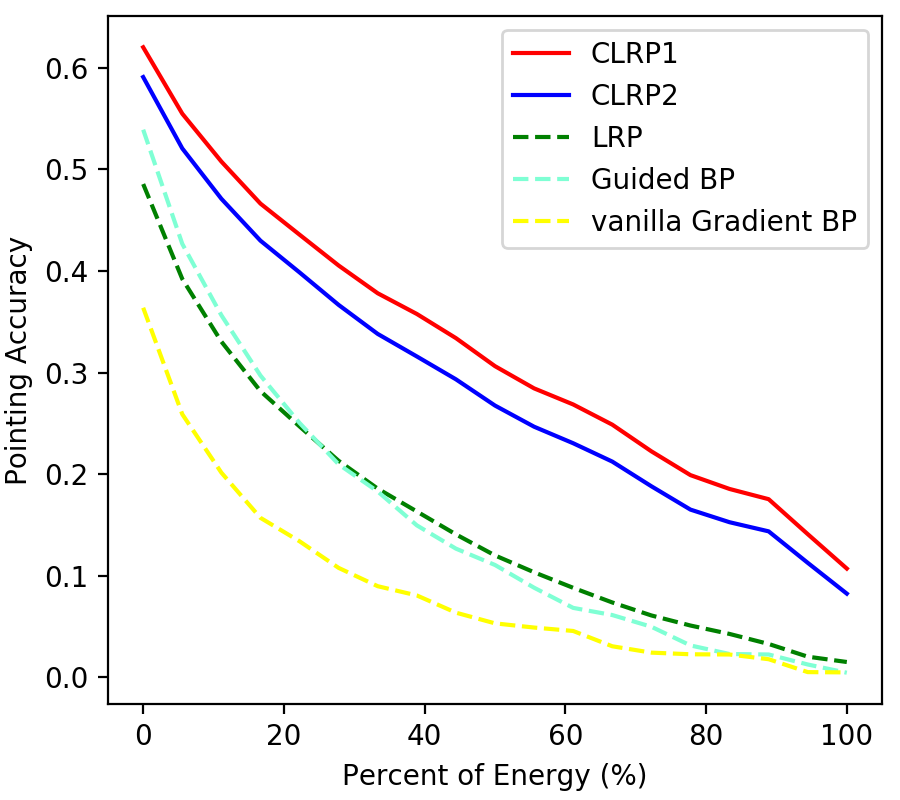}
        \caption{Pointing Accuracy On the AlexNet}
    \end{subfigure}
     \begin{subfigure}[t]{0.45\textwidth}
        \centering
        \includegraphics[scale=0.41]{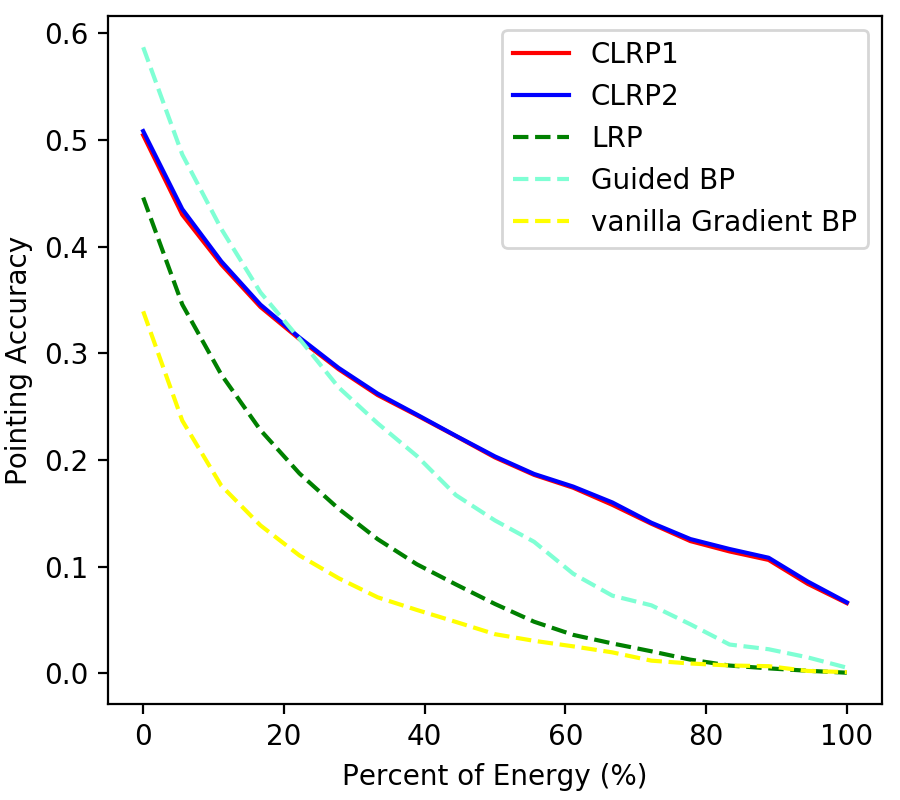}
        \caption{Pointing Accuracy On the VGG16}
    \end{subfigure}
    \caption{The figure shows the localization ability of the saliency maps generated by the LRP, the CLRP1, the CLRP2, the vanilla Gradient Visualization and the Guided Backpropagation. On the pre-trained models, AlexNet and VGG16, the localization ability is evaluated at different thresholds. The x-axis corresponds to the threshold that keeps a certain percentage of energy left, and the y-axis corresponds to the pointing accuracy.}
\label{fig:pointing}
\end{figure}

\textbf{Pointing Game:} To evaluate the discriminativeness of saliency maps, the paper \cite{zhang2016top} proposes a pointing game. The maximum point on the saliency map is extracted and evaluated. In case of images with a single object, a hit is counted if the maximum point lies in the bounding box of the target object, otherwise a miss is counted. The localization accuracy is measured by $Acc = \frac{\#Hits}{\#Hits+\#Misses}$. In case of ILSVRC2012 dataset, the naive pointing at the center of the image shows surprisingly high accuracy. Based on the reason, we extend the pointing game into a difficult setting. In the new setting, the first step is to preprocess the saliency map by simply thresholding so that the foreground area covers $p$ percent energy out of the whole saliency map (where the energy is the sum of all pixel values in saliency map). A hit is counted if the remaining foreground area lies in the bounding box of the target object, otherwise a miss is counted. 

The figure \ref{fig:pointing} show that the localization accuracy of different approaches in case of different thresholds. With more energy kept, the remained pixels are less likely to fall into the ground-truth bounding box, and the localization accuracy is low correspondingly. The CLRP1 and the CLRP2 show constantly much better pointing accuracy than that of the LRP. The positive results indicate that the pixels that the contrastive backpropagation cancels are on the cluttered background or non-target objects. The CLRP can focus on the class-discriminative part, which improves the LRP.  The CLRP is also better than other primal backpropagation-based approaches. One exception is that the Guided Backpropagation shows a better localization accuracy in VGG16 network in case of high thresholds. In addition, the localization accuracy of the CLRP1 and the CLRP2 is similar in the deep VGG16 network, which indicates the equivalence of the two methods to model the opposite visual concept.

\begin{table}[t]
\begin{center}
\caption{Ablation study on ImagNet Validation dataset. The dropped activation values after the corresponding ablation are shown in the table.}
\label{tab:ablation_study}
\def\arraystretch{1.1}
\begin{tabular}{c|c|c|c|c|c|c}
\hline
&Random & vanilGrad\cite{simonyan2013deep} & GuidedBP\cite{springenberg2014striving} & \textbf{LRP}\cite{bach2015pixel} & \textbf{CLRP1} & \textbf{CLRP2}\\
\hline
AlexNet & 0.0766 & 0.1716 & 0.1843 &\quad 0.1624 \quad & \quad 0.2093 \quad & \quad 0.2030 \quad \\
\hline
VGG16 & 0.0809 & 0.3760 & 0.4480 &\quad 0.3713 \quad & \quad 0.3844 \quad  & \quad 0.3913 \quad \\
\hline
\end{tabular}
\end{center}
\end{table}
\textbf{Ablation Study:} In the Pointing Game above, we evaluate the discriminativeness of the explanations according to the localization ability. In this ablation study, we evaluate the discriminativeness from another perspective. We observe the changes of activation in case of ablating the found discriminative pixels. The activation value of the class-specific neuron will drop if the ablated pixels are discriminative to the corresponding class.

For an individual image classification decision, we first generate a saliency map for the ground-truth class. We identify the maximum point in the generated saliency map as the most discriminative position. Then, we ablate the pixel of the input image at the identified position with a $9\times9$ image patch. The pixel values of the image patch are the mean value of all the pixel values at the same position across the whole dataset. We classify the perturbated image and observe the activation value of the neuron corresponding to the ground-truth class. The dropped activation value is computed as the difference between the activations of the neuron before and after the perturbation. The dropped score is averaged on all the images in the dataset.

The experimental results of different approaches are shown in the table \ref{tab:ablation_study}. For the comparison, we also ablate the image with a randomly chosen position. The random ablation has hardly impact on the output. The saliency maps corresponding to all other approaches find the relevant pixel because the activations of the class-specific neurons dropped a lot after the corresponding ablation. In both networks, CLRP1 and CLRP2 show the better scores, which means the discriminativeness of explanations generated by CLRP is better than that of the LRP. 
Again, the Guided Backpropagation shows better score than CLRP. This ablation study only considers the discriminative of the pixel with maximal relevance value, which corresponds to a special case in the Pointing Game, namely, only one pixel with maximal relevance is left after the thresholding. The two experiments show the consistent result that the Guided Backpropagation is better than LRP in the special case. We do not report the performance of the GoogLeNet in the experiments. Our approach shows that the zero-padding operations of convolutional layers have a big impact on the output of the GoogLeNet model in $torchvision$ module of Pytorch. The impact leads to a problematic saliency map (see supplementary material).

\subsection{Understanding the Difference between Neurons}
The neurons of DNNs have been studying with their activation values. The DeConvNets \cite{zeiler2014visualizing} visualize the patterns and collect the images that maximally activate the neurons, given an image set. The activation maximization method \cite{erhan2009visualizing,nguyen2016synthesizing} aims to generate an image in input space that maximally activates a single neuron or a group of neurons. Furthermore,  the work \cite{zhou2014object,gonzalez2018semantic} understand the semantic concepts of the neurons with an annotated dataset. In this experiment, we aim to study the difference among neurons in a single classification decision.

The neurons of low layers may have different local receptive fields. The difference between them could be caused by the different input stimuli. We visualize high-level concepts learned by the neurons that have the same receptive fields, e.g., a single neuron in a fully connected layer. For a single test image, the LRP and the CLRP2 are applied to visualize the stimuli that activate a specific neuron. We do not use CLRP1 because the opposite visual concept cannot be modeled by the remaining neurons in the same layer.

\begin{figure}[t]
        \centering
       \begin{subfigure}{0.33\textwidth}
        \centering
        \includegraphics[scale=0.17]{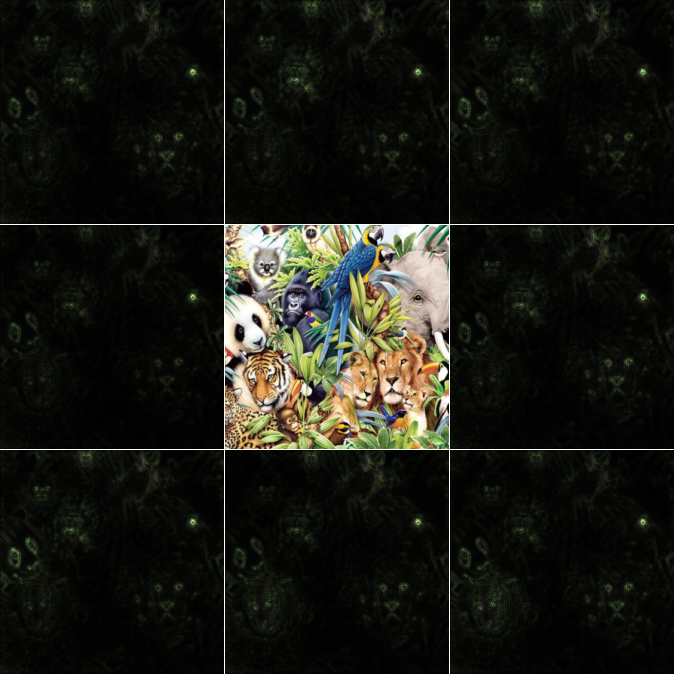}
        \caption{Explanations by LRP}
        \label{fig:vgg_fc1_LRP}
     \end{subfigure}
     \begin{subfigure}{0.33\textwidth}
        \centering
        \includegraphics[scale=0.17]{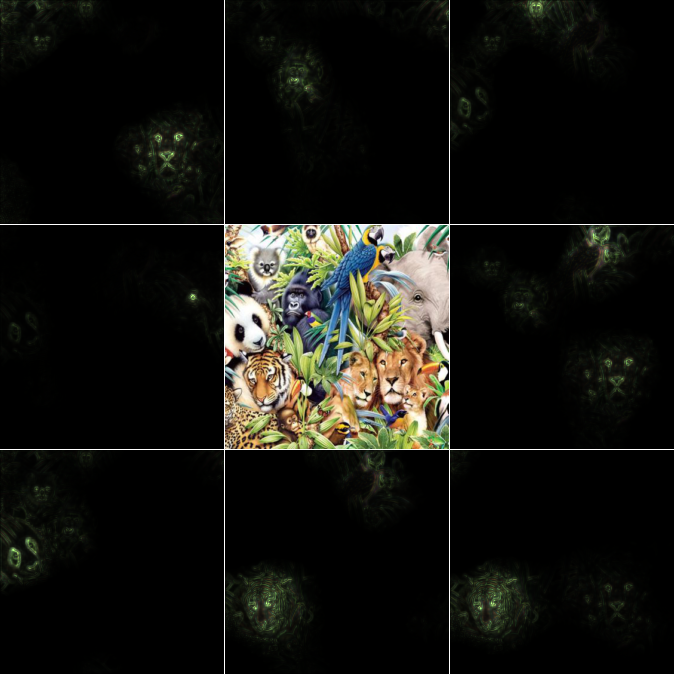}
        \caption{Explanations by CLRP}
        \label{fig:vgg_fc1_CLRP}
    \end{subfigure}
    \begin{subfigure}{0.32\textwidth}
        \centering
        \includegraphics[scale=0.3]{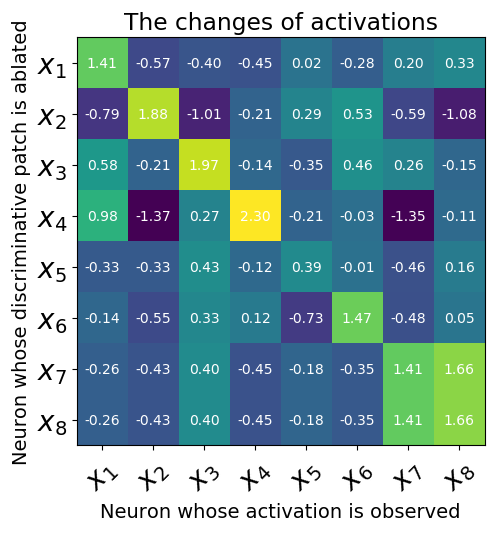}
        \caption{Ablation Study}
        \label{fig:ablation}
    \end{subfigure}
        \caption{The figures show explanation maps of neurons in $fc1$ layers. The explanations generated by LRP are not discriminative. By contrast, the ones generated by CLRP explain the difference between the neurons.}
\label{fig:vgg_neurons}
\end{figure}

In VGG16 network, we visualize 8 activated neurons $x_{1-8}$ from the $fc1$ layer. The visualized maps are shown in figure \ref{fig:vgg_neurons}. The image is classified as a $toyshop$ by the VGG16 network. The receptive field (the input image) is shown in the center, and the 8 explanation maps are shown around it. While the LRP produces almost identical saliency map for the 8 neurons (in figure \ref{fig:vgg_fc1_LRP}), the CLRP2 gains a meaningful insight about their difference, which shows that different neurons focus on different parts of images. By comparison (see figure \ref{fig:vgg_fc1_CLRP}), the neurons $x_1, x_2, x_3$ in the first row are activated more by the $lion$, the $gorilla$, and the $monkey$ respectively. The neurons $x_4, x_5$ in the second row by the eye of the $elephant$ and the $bird$ respectively. The right-down one $x_6$ by the $panda$. The last two neurons $x_7$ and $x_8$ focus on the similar patterns (i.e., the $tiger$).

To our knowledge, there is no known work on the difference between neurons in an individual classification decision and also no evaluation metric. We evaluate the found difference by an ablation study. More concretely, we first find the discriminative patch for each neuron (e.g., $x_{1-8}$) using CLRP2. Then, we ablate the patch and observe the changes of neuron activations in the forward pass. The discriminative patch of a neuron is identified by the point with maximal value in its explanation map created by CLRP2. The $9\times9$ neighboring pixels around the maximum point are replaced with the values that are mean of pixel values in the same positions across the whole dataset.

The ablation study results are shown in the figure \ref{fig:ablation}. The positive value in the grid of the figure means the decreased activation value, and the negative ones mean the activations increase after the corresponding ablation. In case of the ablation corresponding to neuron $x_i$, we see that the activation of $x_i$ is significantly dropped (could become not-activated). The maximal droped values of each row often occur on the diagonal axis. We also try with other ablation sizes and other neurons, which shows the similar results. The ablations for the last two neurons $x_7$ and $x_8$ are same because their explanation maps are similar. The changes of activations of all other neurons are also the same for the same ablation. We found that many activated neurons correspond to same explanation maps.

\section{Conclusion}
The explanations generated by LRP are evaluated. We find that the explanations are not class-discriminative. We discuss the theoretical foundation and provide our justification for the non-discriminativeness. To improve discriminativeness of the generated explanations, we propose the Contrastive Layer-wise Relevance Propagation. The qualitative and quantitative evaluations confirm that the CLRP is better than the LRP. We also use the CLRP to shed light on the role of neurons in DCNNs.

We propose two ways to model the opposite visual concept the class-specific neuron represents. However, there could be other more appropriate modeling methods. Even though our approach produces a pixel-wise explanation for the individual classification decisions, the explanations for similar classes are similar. The fine-grained discriminativeness are needed to explain the classifications of the intra-classes. We leave the further exploration in future work.

%
%
%
 \bibliographystyle{splncs04}
 \bibliography{mybib}
\end{document}